\documentclass[letterpaper]{article} 
\usepackage{aaai2026}  
\usepackage{times}  
\usepackage{helvet}  
\usepackage{courier}  
\usepackage[hyphens]{url}  
\usepackage{graphicx} 
\usepackage{natbib}  
\usepackage{caption} 
\usepackage{algorithm}
\usepackage{algorithmic}
\usepackage{bm}
\usepackage[dvipsnames,table]{xcolor}
\usepackage{multirow}
\usepackage{amsmath}
\usepackage{amssymb}

\usepackage{newfloat}
\usepackage{listings}
\DeclareCaptionStyle{ruled}{labelfont=normalfont,labelsep=colon,strut=off} 
\floatstyle{ruled}
\newfloat{listing}{tb}{lst}{}
\floatname{listing}{Listing}
\title{Continual Learning for Multiple Modalities}

\author{Hyundong Jin and Eunwoo Kim\\
School of Computer Science and Engineering\\
Chung-Ang University, South Korea\\
{\tt\small jude0316@cau.ac.kr ~ eunwoo@cau.ac.kr}
}

\begin{document}

\maketitle

\begin{abstract}
Continual learning aims to learn knowledge of tasks observed in sequential time steps while mitigating the forgetting of previously learned knowledge.
Existing methods were designed to learn a single modality (e.g., image) over time, which limits their applicability in scenarios involving multiple modalities.
In this work, we propose a novel continual learning framework that accommodates multiple modalities (\textit{image, video, audio, depth, and text}).
We train a model to align various modalities with text, leveraging its rich semantic information.
However, this increases the risk of forgetting previously learned knowledge, exacerbated by the differing input traits across tasks.
To alleviate the overwriting of previous knowledge of modalities, we propose a framework that consolidates intra-modal knowledge while incorporating relevant inter-modal information. 
This is achieved by self‑regulating shifts in learned representations to gradually integrating novel knowledge into the information retained across modalities.
Simultaneously, it mitigates inter-modal interference by selectively integrating knowledge from previously encountered modalities based on their mutual relevance.
Furthermore, we introduce a strategy to re-align modality embeddings, effectively addressing biased alignment between modalities.
We evaluate the proposed method in a wide range of continual learning scenarios using multiple datasets with different modalities.
Extensive experiments demonstrate that ours outperforms existing methods in the scenarios, regardless of whether the identity of the modality is given.
\end{abstract}



\section{Introduction}
\label{sec:introduction}
Recently, a vision-language model, pre-trained with a large number of image-text pairs, has drawn significant attention \cite{radford2021learning} due to its strong generalization performance \cite{lin2023clip, rasheed2023fine}.
Leveraging the general knowledge of the pre-trained model, several algorithms have achieved promising results in image classification \cite{khattak2023maple}, semantic segmentation \cite{luddecke2022image}, and video action recognition \cite{rasheed2023fine}, to name a few.
However, when deep learning models are exposed to new data, they tend to overwrite previously learned knowledge, leading to forgetting earlier information.
This issue is known as catastrophic forgetting \cite{mccloskey1989catastrophic, kirkpatrick2017overcoming}.

\begin{figure}[t!] 
    \centering \includegraphics[width=\columnwidth]{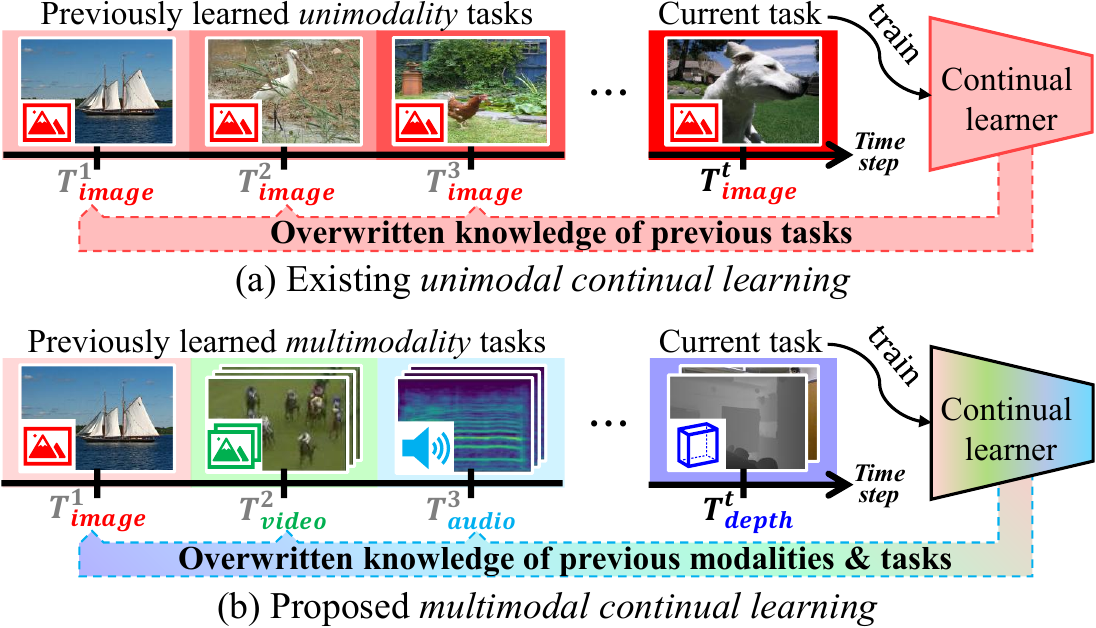}
    \caption{Illustration of (a) existing unimodal and (b) the proposed multimodal continual learning.
    Multimodal continual learning poses a significant challenge, as it complicates the retention of previously acquired knowledge while integrating information from tasks across diverse modalities.
    }
    \label{fig: unimodal versus multimodal}
\end{figure}

To alleviate this phenomenon, continual learning methods strive to preserve existing knowledge while accommodating new information \cite{kirkpatrick2017overcoming, aljundi2018memory, rebuffi2017icarl, smith2023coda, hou2018lifelong, li2016learning}.
To this end, they apply penalties to minimize changes in key parameters that contribute to performance on previous tasks \cite{kirkpatrick2017overcoming, aljundi2018memory} or distill the knowledge of the old model into newer ones \cite{hou2018lifelong, li2016learning}.
Furthermore, to facilitate recall of previous knowledge, some methods rely on having access to data from old tasks \cite{ostapenko2019learning, rebuffi2017icarl, douillard2022dytox, Liu2020AANets}. 
The collected data is then used jointly with new data to train the model. 
However, storing raw data can lead to privacy concerns in many applications \cite{smith2023closer}. 
Moreover, temporal data (e.g., video and audio) demand more memory resources than non-temporal data \cite{villa2022vclimb, jin2023growing}, posing additional challenges for continual learning.

Recently, continual learning methods that adapt a pre-trained model (e.g., ViT \cite{dosovitskiy2020image} or CLIP \cite{radford2021learning}) have been proposed \cite{wang2022learning, wang2023attriclip, he2025cl}.
These methods are designed to select prompts from a shared pool for each task.
The selected prompts are inserted into a pre-trained model and updated to acquire new knowledge.
Despite outperforming other methods by leveraging the rich knowledge of the pre-trained model, these approaches validate on a single modality and do not account for data from new modalities.
This limitation restricts their applicability in dynamic environments where data from different modalities appear, such as in autonomous driving \cite{prakash2021multi} and robot planning \cite{huang2025thinkact}.
More importantly, these approaches are susceptible to forgetting knowledge learned previously when learning multiple modalities because modality gaps between tasks cause prompts trained on different input types to interfere with each other.

In this paper, we propose a novel approach, \textbf{COMM} (\textbf{CO}ntinual Learning for \textbf{M}ultiple \textbf{M}odalities), for continuously learning tasks of different modalities, using a pre-trained vision-language model.
Specifically, we take text-bound data as input, leveraging the rich semantic information provided by text \cite{zhu2023languagebind}.
We process the text data through a language encoder and the other modalities through a modality encoder, enabling the integration of additional modalities without introducing many parameters.
To leverage rich pre-trained knowledge, we introduce prompts to accommodate new knowledge without updating the backbone parameters.
These prompts are accumulated over time for each modality through self-regularization, which preserves representational consistency across time and supports stable integration of intra-modality knowledge.
To reduce interference across modalities, we promote modality-aware adaptation by aggregating prompts from previously encountered modalities, suppressing the influence of unrelated ones.
Additionally, to prevent biased alignment of recently learned text \cite{wu2019large}, we re-align earlier modality representations to restore semantic consistency with their textual counterparts.
Figure \ref{fig: unimodal versus multimodal} illustrates the proposed \textit{multimodal continual learning} compared to existing unimodal continual learning.

We demonstrate our method by comparing it with existing continual learning methods across tasks involving multiple modalities using classification benchmark datasets: ImageNet-100 \cite{deng2009imagenet}, UCF-101 \cite{soomro2012ucf101}, SUN-RGBD \cite{song2015sun}, and ESC-50 \cite{piczak2015esc}.
Experimental results show that the proposed method outperforms its competitors across various class-incremental learning scenarios while introducing negligible memory overhead from the prompts.
The contributions of our work are four-fold: 
\begin{itemize}
\item We propose the first multimodal continual learning method that leverages multiple modalities, addressing the challenge of forgetting that arises from learning the distinct traits of each modality in sequential data.

\item We present a knowledge aggregation approach that effectively learns a new task by mitigating interference between modalities while preserving existing knowledge.

\item Our re-alignment strategy restores distorted semantic consistency between previously learned modalities and text during adaptation to new tasks.

\item Extensive experiments show that the proposed method achieves outstanding performance compared to existing continual learning methods by effectively preventing detrimental knowledge mixing across modalities.
\end{itemize}

\section{Related Work}
\noindent{\textbf{Multimodal learning}} aims to construct unified representations from diverse data types, with significant progress driven by 
pre-trained models such as CLIP \cite{radford2021learning}.
These models are commonly adapted to downstream tasks, including image classification \cite{khattak2023maple, zhou2022learning, wortsman2022robust} and video-text matching \cite{lin2022frozen, rasheed2023fine}, via fine-tuning pre-trained weights or inserting learnable prompts.
Recent studies \cite{girdhar2023imagebind, zhu2023languagebind, zhou2025unialign} have extended this framework to encompass three or more modalities, including image and text.
While these approaches train modalities jointly, sequential training poses a greater challenge, as it must preserve representations of previously learned modalities while integrating new ones.
Moreover, existing methods often rely on modality-specific networks \cite{girdhar2023imagebind, zhu2023languagebind}, limiting scalability as the number of modalities grows.
In this work, we present a global encoder for non-text modalities that aligns prompts across modalities, mitigating interference and enhancing representational consensus.

\noindent{\textbf{Continual learning}} aims to maintain previously learned knowledge while simultaneously acquiring new information. 
Existing continual learning methods have primarily focused on a single modality (e.g., image).
Early works proposed to identify contributing parameters for previous tasks \cite{kirkpatrick2017overcoming, zenke2017continual, aljundi2018memory}.
These methods train the model by penalizing updates to key parameters, which alleviates forgetting of previously learned knowledge.
However, they can cause significant loss of knowledge as the number of tasks with different semantic domains increases \cite{de2021continual}.
Several studies store \cite{douillard2022dytox, Liu2020AANets, park2021class} or generate \cite{ostapenko2019learning, shin2017continual, wang2019continual} some data from old tasks and train the network with a joint set of old and new data. 
However, privacy or memory budget issues can arise from storing old data \cite{villa2022vclimb}.
Importantly, the unimodal focus of these approaches limits their applicability in real-world settings where data arrives as a continuous stream from multiple modalities.

\begin{figure*}[t!] 
    \centering 
    \includegraphics[width=0.97\textwidth]{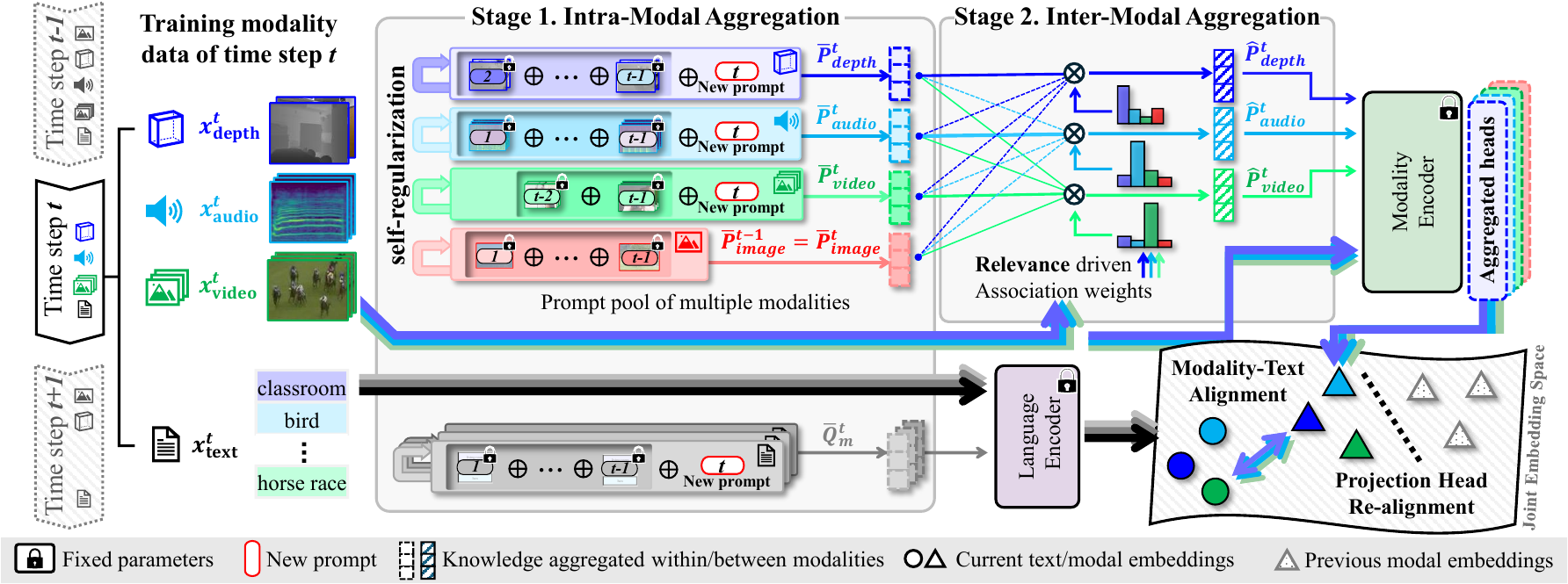}
    \caption{An illustration of the proposed method for learning multiple modalities at time step $t$.
    The model receives modality data paired with the corresponding text.
    First, new prompts are introduced and aggregated with previously accumulated ones via self-regularization to preserve intra-modal knowledge.
    Then, intra-modal prompts are integrated into final prompts using modality-aware relevance weights, which suppress cross-modal interference.
    The resulting prompts are passed into the modality encoder to obtain aligned features.
    To prevent biased projections toward recently learned texts, projection heads are re-aligned to restore semantic consistency with previously learned alignments.
    }
    \vspace{-2mm}
    \label{fig: COMM overview} 
\end{figure*}

\noindent{\textbf{Continual learning with a vision-language model}} \cite{radford2021learning} has recently been explored, with methods adapting to new unimodal data by matching it with corresponding text descriptions \cite{wortsman2022robust, villa2023pivot, wang2023attriclip, wu2025synthetic}.
Some methods \cite{zheng2023preventing, wortsman2022robust} directly update the parameters of a model while regularizing to retain its learned information.
The others \cite{villa2023pivot, wang2023attriclip, qiao2024PGP} learn new tasks by combining trainable prompts with the input embeddings in each layer of a pre-trained model, selecting prompts from a predefined pool.
However, these methods often overwrite previously learned knowledge by updating model parameters or prompts, exacerbating forgetting due to interference from different modalities across tasks.
In contrast, we aggregate prompts from modalities relevant to the input, ensuring that knowledge learned from other modalities remains unchanged, alleviating interference between modalities.

\section{Methodology}

\subsection{Problem Definition}
We present a new problem of learning multiple modalities over a sequence of tasks, where at each time step $t$, a subset of modalities, $\mathcal{M}^{t} \subseteq \mathcal{M} = \{$image, video, depth, audio$, ...\}$, is presented.
Specifically, at time step $t$, we learn knowledge of each modality $m \in \mathcal{M}^{t}$ by aligning it with the corresponding class name (represented by text) in a joint embedding space.
The goal of multimodal continual learning is to maintain previously learned knowledge (modality-text alignments) while integrating new knowledge from the current task.
At time step $t$, the task\footnote{We assume that a task appears at every time step.} is defined by a set of modality-text pairs $T^{t}=\{D^{t}_{m}, C^{t}_{text}\}_{m\in\mathcal{M}^{t}}$.
For each modality $m$, $D^{t}_{m}$ comprises data points and their labels $\{x^{t}_{m, i}, y^{t}_{i}\}_{i=1}^{n^{t}_{m}}$, and $C^{t}_{text}$ provides the corresponding textual class names $\{x^{t}_{\text{text},i}\}_{i=1}^{n^{t}_{m}}$.

\noindent\textbf{Modality-text alignment.} To extract embeddings of the modality data $x^{t}_{m,i}$ and the text data $x^{t}_{\text{text},i}$, we use the modality\footnote{We implement the modality encoder using the vision encoder in \cite{radford2021learning}.} and language encoders, $V(\cdot)$ and $L(\cdot)$, in a pre-trained model \cite{radford2021learning}, respectively. 
This approach helps align embeddings from non-text modalities with those from text \cite{girdhar2023imagebind, zhu2023languagebind}.
To mitigate the increase in memory cost as the number of modalities grows, non-text modalities share a common modality encoder.
We obtain the modality feature $\tilde{v}^{t}_{m,i}=V(x^{t}_{m,i})$ and the text embedding $l^{t}_{\text{text}, i}=L(x^{t}_{\text{text},i})$, respectively.
$\tilde{v}^{t}_{m,i}$ is further projected to a common modal-text embedding space as $v^{t}_{m,i} = f_{\alpha^{t}_{m}}(\tilde{v}^{t}_{m,i})$, where $f_{\alpha^{t}_{m}}(\cdot)$ is a projection head parameterized by $\alpha^{t}_{m}$.
To learn the modality data, we maximize the predictive probability defined as
\begin{equation}
    \label{equation: probability}
    p(y^{t}_{i}|x^{t}_{m,i}) = \frac{\mbox{exp}(\mbox{sim}(v^{t}_{m,i}, l^{t}_{\text{text}, i}) / \tau )}
    {\sum_{j} \mbox{exp}(\mbox{sim}(v^{t}_{m,i}, l^{t}_{\text{text}, j}) / \tau ) },
\end{equation}
where $\mbox{sim}(\cdot)$ and $\tau$ represent the cosine similarity and the temperature parameter, respectively.

\subsection{Framework}\label{sec:3.2}
The proposed method learns multimodal tasks based on a pre-trained model.
However, directly updating the model on different modalities can exacerbate forgetting due to interference caused by modality gaps; representational discrepancies and conflicting updates from informational imbalance \cite{peng2022balanced}.
To resolve it, we take a strategy of introducing prompts to learn new tasks without updating the model itself \cite{khattak2023maple}.
Figure \ref{fig: COMM overview} illustrates our continual learning method for multiple modalities.

Specifically, the set of modalities of the task at time $t$ is $\mathcal{M}^{t}$, where each modality consists of text-paired data.
To learn new tasks, we introduce and update a pair of learnable prompts, $P^{t}_{m}$ and $Q^{t}_{m}$, of the modality and language encoders by incorporating them with the input, respectively.
Unlike existing prompt-based continual learning \cite{zhou2022learning, wang2023attriclip}, which employs a prompt pool for unimodal tasks, multimodal continual learning requires a shared pool capable of accommodating tasks with diverse modalities.
This introduces a challenge in selecting appropriate prompts.

To effectively utilize prompts for each modality and minimize interference with previously learned representations, we propose aggregating prompts intra and inter modalities in the proposed pipeline.
In the first stage, to alleviate significant changes in old knowledge across time within a modality, we combine the prompts learned so far with newly introduced ones by a self-regulating strategy.
The update of prompts across modalities can lead to knowledge degradation due to input discrepancies; thus, the second stage mitigates forgetting by aligning knowledge through modality-aware associations.
We obtain modality and text embeddings using prompts aggregated intra and inter modalities.
The embeddings are aligned by updating $P^{t}_{m}$ and $Q^{t}_{m}$ using the cross-entropy loss, $\mathcal{L}_{\text{task}} = - \sum_{i} y^{t}_{i} \log p(y^{t}_{i}|x^{t}_{m,i}).$

\subsection{Continual Learning for Multiple Modalities}
\noindent\textbf{Intra-modal aggregation via self-regularization.}
To maintain earlier representations and ensure effective prompt utilization within each modality, we aggregate prompts over time and self-regulate them.
Specifically, prompts $P^{t}_{m}$ and $Q^{t}_{m}$ are zero-initialized to gradually acquire knowledge of the task.
These prompts are then integrated with those from the previous tasks as $\bm{\bar}{P}^{t}_{m} = \oplus(\bm{\bar}{P}^{t-1}_{m},P^{t}_{m})$ and $\bm{\bar}{Q}^{t}_{m} = \oplus(\bm{\bar}{Q}^{t-1}_{m},Q^{t}_{m}$), 
where $\bm{\bar}{P}^{1}_m = P^{1}_m$ and $\bm{\bar}{Q}^{1}_m = Q^{1}_m$. 
$\oplus$ denotes element-wise summation, and prompts absent in earlier time steps are excluded from this operation.
The aggregated prompts are used to extract modality-text embeddings, 
$V_{\bm{\bar}{P}^{t}_{m}}(x^{t}_{m,i})$ and $L_{\bm{\bar}{Q}^{t}_{m}}(x^{t}_{\text{text},i})$,
which are aligned by minimizing the loss, $\mathcal{L}_{\text{task}}$.
However, optimizing solely for the task loss may fail to adequately preserve prior knowledge, leading to significant shifts in the accumulated representations.

To address this, we introduce a self-regularization loss, $\mathcal{L}_{\text{self}}$, applied to both the prompts and the projection heads, enforcing consistency with previously accumulated knowledge.
Projection head parameters are similarly aggregated over time: $\bm{\bar}{\alpha}^{t}_{m}$= $\oplus(\bm{\bar}{\alpha}^{t-1}_{m}, \alpha^{t}_{m})$.
The complete self-regularization loss is then defined as 
\begin{equation}
    \mathcal{L}_{\text{self}} = 
    \sum_{i}  \langle V_{\bm{\bar}{P}^{t}_{m}}, x^{t}_{m,i} \rangle + \langle L_{\bm{\bar}{Q}^{t}_{m}}, x^{t}_{\text{text},i} \rangle + \langle f_{\bm{\bar}{\alpha}^{t}_{m}}, \bar{v}_{m,i}^t \rangle, 
\end{equation}
where 
$\langle V_{A_m^{t}}, b \rangle = \Vert V_{A_m^{t}}(b) - V_{A_m^{t-1}}(b) \Vert_2$ denotes the $\ell_2$ distance between the outputs of $V_{A_m}(\cdot)$ given an input $b$ using parameters at the current and previous time steps, and $\bar{v}_{m,i}^t = V_{\bm{\bar}{P}^{t}_{m}}(x^{t}_{m,i})$.
This loss penalizes output changes of the prompts and projection head by comparing their current outputs with those obtained using the previously aggregated parameters.
Incorporating $\mathcal{L}_{\text{self}}$ into $\mathcal{L}_{\text{task}}$ encourages stable knowledge integration into the parameters $\bm{\bar}{P}^{t}_{m}, \bm{\bar}{Q}^{t}_{m}$, and $\bm{\bar}{\alpha}^{t}_{m}$. 
This strategy enables gradual adaptation of prompts and projection heads while it preserves prior knowledge and prevents interference from incorrect prompt selection.

\noindent\textbf{Inter-modal integration guided by relevance.}
Prompts trained across modalities are more vulnerable to interference when applied in unrelated contexts than those trained for a single modality.
This necessitates the identification of relevant modalities from the pool observed thus far, denoted as $\mathcal{M}^{1:t}$.
To this end, we introduce a function $g_{\beta^{t}}(\cdot)$: $\mathbb{R}^{d_{i}} \rightarrow \mathbb{R}^{|\mathcal{M}^{1:t}|}$, parameterized with $\beta^{t}$, to identify relevant modalities by examining the features extracted from the modality encoder, $V(\cdot)$.
The function assigns probabilities to the prompts, representing their relevance and facilitating the integration of prompts from the modalities $\{\bm{\bar}{P}^{t}_{m'}\}_{m'\in\mathcal{M}^{1:t}}$.
Due to the inaccessibility of old data, we collect features for each modality, which are sampled from a normal distribution defined by the mean and covariance of the previously observed features \cite{zhang2023slca}.
The function $g_{\beta^{t}}(\cdot)$ is trained by minimizing 
\begin{equation}
\mathcal{L}_{\text{cross}} = 
        \sum_{m' \in \mathcal{M}^{1:t}}
        \sum_{i=1}^{\bar{n}_{m'}}
        ~ -\log g_{\beta^{t}}(m'|\bar{u}_{m',i}),
\end{equation}
where $\bar{u}_{m',i}$ is a collected sample and $\bar{n}_{m'}$ denotes the number of samples.

By minimizing this loss, $g_{\beta^{t}}$ yields relevance of each modality $m'\in \mathcal{M}^{1:t}$ in relation to $m$.
The prompts are then aggregated across modalities using the relevance $g_{\beta^{t}}$ as weights:
\begin{equation}
    \bm{\hat}{P}^{t}_{m} = \sum_{m' \in \mathcal{M}^{1:t}} 
        g_{\beta^{t}}(m'|V(x^{t}_{m,i})) \cdot \bm{\bar}{P}^{t}_{m'},
\end{equation}
This composed prompt $\bm{\hat}{P}^{t}_{m}$ selectively emphasizes prompts associated with modality contexts relevant to the current input, while attenuating the influence of unrelated ones.
Note that the text prompts do not require such aggregation because text data is input solely into the language encoder.
By aligning the modality and text embeddings extracted using the prompts, $\bm{\hat}{P}^{t}_{m}$ and $\bm{\bar}{Q}^{t}_{m}$, we can avoid interference between modalities.
To summarize, we update the set of parameters $\{P^{t}_{m}, Q^{t}_{m}, \alpha^{t}_{m}\}$ by minimizing the aforementioned loss functions.

\noindent\textbf{Modality-text re-alignment.}
\label{section:3.5 Head alignment}
Learning new knowledge without accessing data from old tasks can cause the embeddings from the modality encoder to be biased toward the text embeddings of the latest task \cite{yu2020semantic}, which distorts the previously modality-text alignments. 
This degrades the semantic grounding of accumulated prompts, as they are composed based on earlier alignments, making the  modality-text inconsistency more severe.
To address this issue, we introduce an additional phase to re-align the projection head, $f_{\bm{\bar}{\alpha}^{t}_{m}}(\cdot)$, so that old and new modality embeddings are distinguished.
Specifically, for each class of modality $m$, we sample features that mimic old ones using the class-wise mean and covariance of previously stored embeddings, which are extracted via the modality encoder with the prompt, $\bm{\hat}{P}^{i}_{m}$, where $i<t$.
The aggregated projection head, $\bm{\bar}{\alpha}^{t}_{m}$, is further fine-tuned to align the modality embedding derived from the sampled features with the text embedding by minimizing $\mathcal{L}_{\text{task}}$.

\renewcommand{\arraystretch}{1.0}
\setlength{\tabcolsep}{3pt}
\begin{table*}[t]
\centering
\resizebox{0.97\textwidth}{!}{%
\small
\setlength\doublerulesep{.5pt}
\begin{tabular}{l||ccc|ccc|ccc|ccc|ccc}
\hline
\multirow{2}{*}{Method} 
& \multicolumn{3}{c|}{Image}
& \multicolumn{3}{c|}{Video} 
& \multicolumn{3}{c|}{Depth} 
& \multicolumn{3}{c|}{Audio} 
& \multicolumn{3}{c}{Overall}\\ \cline{2-16} 
&AIA ($\uparrow$) &FAA ($\uparrow$) & F ($\downarrow$)
&AIA ($\uparrow$) &FAA ($\uparrow$) & F ($\downarrow$)
&AIA ($\uparrow$) &FAA ($\uparrow$) & F ($\downarrow$)
&AIA ($\uparrow$) &FAA ($\uparrow$) & F ($\downarrow$)
& AIA ($\uparrow$) & FAA ($\uparrow$) & F ($\downarrow$) \\ \hline
\multicolumn{16}{c}{\textbf{Modality-Specific Class-incremental Learning}} \\ \hline
FT 
& 45.66    & 23.46 & 23.37
& 45.25    & 13.16 & 33.87        
& 26.21    & 7.58  & 19.72            
& 28.96    & 4.20  & 26.31          
& 36.52    & 12.16 &  25.82 \\
EWC \cite{kirkpatrick2017overcoming}
& 47.16    & 23.42   & 24.98   
& 48.59    & 20.76   & 29.37     
& 26.38    & 13.91   & 13.20     
& 26.68    & 5.50    & 20.50       
& 37.20    & 15.90   & 22.51 \\
LwF \cite{li2016learning}
& 53.13    & 24.78     & 29.84 
& 48.59    & 23.26     & 26.74
& 30.64    & 11.43     & 20.34   
& 31.89    & 9.75      & 23.53    
& 41.06    & 17.31     & 25.11 \\
WISE-FT \cite{wortsman2022robust}
& 46.73    & 21.96  & 26.08 
& 46.99    & 21.27  & 27.14     
& 26.93    & 5.27   & 22.93     
& 22.69    & 8.25   & 15.34       
& 35.83    & 14.18  & 22.87 \\
L2P \cite{wang2022learning}
& 76.96    & 68.66  & 8.74 
& 78.49    & 70.71  & 8.21     
& 45.06    & 24.75  & 24.66     
& 44.28    & 17.00  & 28.98         
& 61.20    & 45.28  & 17.65 \\
S-liPrompts \cite{wang2022s}
& 75.61    & 65.25  & 10.91 
& 82.85    & 69.04  & 14.57     
& 47.98    & 26.08  & 30.07 
& 42.68    & 22.25  & 21.70 
& 62.28    & 45.66  & 19.31\\
AttriCLIP \cite{wang2023attriclip}
& 77.39    & 65.70  &  12.30 
& 77.40    & 70.51  & 7.28   
& 51.21    & 35.01  & 17.16       
& 43.44    & 20.50  & 24.37
& 62.36    & 47.93  & 15.27 \\ 
PGP \cite{qiao2024PGP}
& 78.88    & 68.58  & 10.84   
& 83.14    & 71.46  & 12.33    
& 48.67    & 25.58  & 24.45      
& 47.69    & 26.25  & 22.78 
& 64.60    & 47.98  & 17.60 \\ 
\rowcolor[gray]{0.9}COMM (Ours)
& \textbf{86.07}    & \textbf{78.96} & 7.11
& \textbf{91.03}    & \textbf{83.82} & 7.21 
& \textbf{62.90}    & \textbf{44.03} & 18.88       
& \textbf{61.86}    & \textbf{43.25} & 18.60 
& \textbf{75.47}   & \textbf{62.51}  & 13.71 \\ \hline \hline
\multicolumn{16}{c}{\textbf{Modality-Agnostic Class-incremental Learning}} \\ \hline
FT 
& 33.39    & 1.46        & 33.61    
& 27.90    & 0.14      & 29.30 
& 25.70    & 7.59    & 19.18   
& 23.65    & 4.25  & 20.61 
& 27.66     & 3.36 & 25.68\\
EWC \cite{kirkpatrick2017overcoming}
& 38.23   & 5.60  & 34.34  
& 27.46   & 3.63  & 25.15 
& 24.04   & 13.91   &  10.73     
& 22.77   & 5.50      & 18.34        
& 28.13  & 7.15   & 22.14 \\
LwF \cite{li2016learning}
& 48.74    & 21.14    & 29.05 
& 19.82    & 1.17     & 22.43 
& 23.01    & 6.74     & 17.22 
& 19.01    & 9.00     & 10.70 
& 27.65   & 9.51      & 19.85  \\
WISE-FT \cite{wortsman2022robust}
& 38.43   & 10.34   & 29.57 
& 24.89   & 0.36    & 25.88 
& 25.94   & 5.27    & 21.67 
& 22.46   & 6.25    & 17.22 
& 27.93    & 5.55   & 23.59  \\
L2P \cite{wang2022learning}
& 79.85    & 69.84  & 10.53 
& 71.25    & 65.14  & 6.44 
& 39.92    & 26.72  & 13.96 
& 33.08    & 21.50    &  12.31 
& 56.03   & 45.80   & 19.31 \\
S-liPrompts \cite{wang2022s}
& 77.36    & 66.72  & 11.20 
& 79.52    & 73.18  & 6.69 
& 40.94    & 30.02  & 11.56 
& 33.63    & 23.50  & 10.79 
& 57.87    & 48.36  & 10.06 \\
AttriCLIP \cite{wang2023attriclip}
& 73.93    & 63.90 & 10.56
& 73.28    & 68.70 & 4.83 
& 39.52    & 28.57 & 11.60 
& 34.99    & 22.50 & 13.27 
& 55.43   & 45.91  & 10.06 \\ 
PGP \cite{qiao2024PGP}
& 81.29    & 70.92   & 10.85
& 77.55    & 67.18   & 10.94 
& 45.61    & 26.25   & 20.49 
& 46.15    & 21.50   & 26.19 
& 62.37    & 41.58   & 17.12 \\ 
\rowcolor[gray]{0.9}COMM (Ours)
& \textbf{83.89}    & \textbf{76.84} & 5.43
& \textbf{90.83}    & \textbf{83.63} & 7.60   
& \textbf{62.78}    & \textbf{43.94} & 19.84
& \textbf{61.85}    & \textbf{43.25} & 19.60
& \textbf{74.84}    & \textbf{61.92} & 13.19   \\ \hline 
\end{tabular}%
}
\caption{
Results for the first sequence on the modality-specific (top) and modality-agnostic (bottom) scenarios. 
The AIA, FAA, and F metrics are measured after training all tasks.
Overall denotes the average AIA and FAA across all modalities.}
\label{tab: Scenario - S1}
\end{table*}
\setlength{\tabcolsep}{6pt}
\vspace{-2mm}
\renewcommand{\arraystretch}{1.0}

\begin{figure*}[] 
     \centering \includegraphics[width=0.97\textwidth]{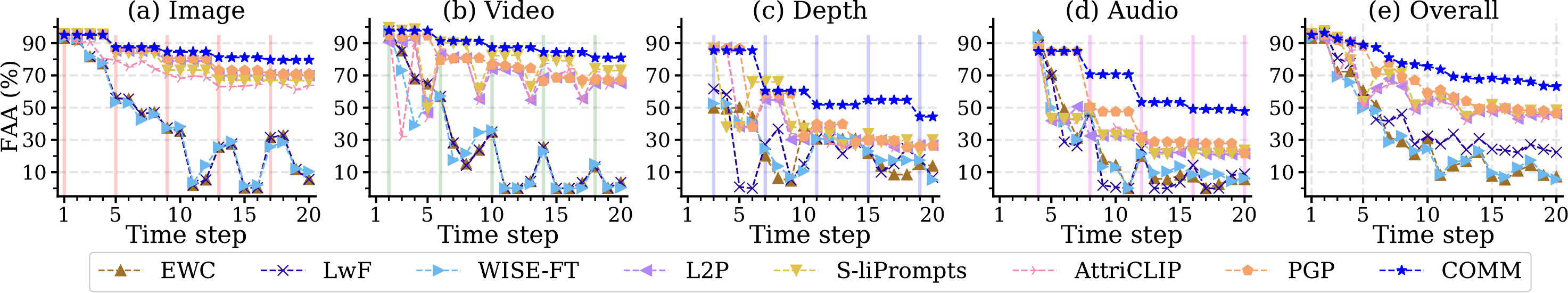}
    \caption{Results of modality-agnostic continual learning for the first sequence, where FAA is measured at each time step.
    The vertical lines represent the time steps at which the modality is learned. 
    }
    \vspace{-2mm}
    \label{fig: S1 modality-agnostic} 
\end{figure*} 

\section{Experiments}
\subsection{Experimental Setup}
\noindent{\textbf{Scenarios.}}
We evaluated the proposed method for continual learning of five modalities: image, video, audio, depth, and text, using four datasets: ImageNet-100 \cite{deng2009imagenet}, UCF-101 \cite{soomro2012ucf101}, ESC-50 \cite{piczak2015esc}, and SUN-RGBD \cite{song2015sun}. 
Each dataset was divided into five non-overlapping subsets of classes.
We consider each subset as a task.
{For a comprehensive evaluation, we designed a benchmark based on} three sequences of tasks: 
(\lowercase\expandafter{\romannumeral1}) A random modality is introduced at each time step. 
(\lowercase\expandafter{\romannumeral2}) A modality is trained across multiple time steps before shifting to the next modality. 
(\lowercase\expandafter{\romannumeral3}) All modalities are available at every time step. 
The first two sequences comprise 20 time steps (corresponding to five subsets of four modalities), with each task exclusively learning a single modality.
In contrast, the third sequence consists of five time steps, during which all modalities are learned simultaneously for every step.
For each sequence, we evaluated the methods with and without the provision of modality identity during the evaluation phase, which we refer to as the \textit{modality-specific} and \textit{modality-agnostic} scenarios, respectively.
We report average incremental accuracy (AIA) \cite{rebuffi2017icarl}, final average accuracy (FAA) \cite{douillard2022dytox}, and average forgetting (F) \cite{chaudhry2019efficient}.

\noindent\textbf{Baselines.}
We evaluated the proposed method, COMM, with existing continual learning methods: conventional approaches, including sequential fine-tuning (FT), EWC \cite{kirkpatrick2017overcoming}, LwF \cite{li2016learning}, and WISE-FT \cite{wortsman2022robust}; and prompt-based methods, L2P \cite{wang2022learning}, S-liPrompts \cite{wang2022s}, AttriCLIP \cite{wang2023attriclip}, and PGP \cite{qiao2024PGP}.

\noindent{\textbf{Implementation details.}}
For a fair comparison with other continual learning methods, we adopted ViT-B-16 CLIP \cite{radford2021learning} as the backbone model for the compared methods.
We used the vision and text encoders of CLIP for the non-text and text modalities, respectively, in accordance with established practices \cite{girdhar2023imagebind, zhu2023languagebind}.
We used a fully connected layer for $g_{\beta^{t}}(\cdot)$.
The proposed method was trained using the Adam optimizer \cite{KingBa15} with $\beta_1$ and $\beta_2$ of 0.9 and 0.999.
Additional information on datasets, preprocessing per modality, baseline setups, and hyperparameters are provided in the supplementary material.

\renewcommand{\arraystretch}{1.0}
\setlength{\tabcolsep}{3pt}
\begin{table*}[t]
\scriptsize
\centering
\small
\setlength\doublerulesep{.5pt}
\resizebox{0.97\textwidth}{!}{%
\begin{tabular}{l||cccccc|cccccc}
\hline
\multirow{3}{*}{Method} & \multicolumn{6}{c|}{\textbf{Modality-Specific Learning}}                       & \multicolumn{6}{c}{\textbf{Modality-Agnostic Learning}}                       \\ \cline{2-13} 
& Image ($\uparrow$) & Video ($\uparrow$) & Depth ($\uparrow$) & \multicolumn{1}{c|}{Audio ($\uparrow$)} & \begin{tabular}[c]{@{}c@{}}Overall\\ AIA ($\uparrow$)\end{tabular} & \begin{tabular}[c]{@{}c@{}}Overall\\ F ($\downarrow$) \end{tabular}
& Image ($\uparrow$) & Video ($\uparrow$) & Depth ($\uparrow$) & \multicolumn{1}{c|}{Audio ($\uparrow$)} & \begin{tabular}[c]{@{}c@{}}Overall\\ AIA ($\uparrow$)\end{tabular}  & \begin{tabular}[c]{@{}c@{}}Overall\\ F ($\downarrow$) \end{tabular} \\ \hline
FT 
& 26.38      & 27.27      & 20.00      & \multicolumn{1}{c|}{26.39}      & 25.01 & 24.56
& 26.05      & 22.61      & 16.45      & \multicolumn{1}{c|}{25.74}      & 22.71 & 22.64       \\
EWC \cite{kirkpatrick2017overcoming}
& 27.21      & 28.45      & 18.19      & \multicolumn{1}{c|}{30.32}      & 26.04  & 24.86      
& 26.10      &  24.37     & 16.98      & \multicolumn{1}{c|}{29.88}      & 24.33  & 26.59      \\
LwF \cite{li2016learning}
& 26.92      & 28.52      & 18.71      & \multicolumn{1}{c|}{27.76}      & 25.48   & 24.83
& 20.48      & 23.12       & 18.23      & \multicolumn{1}{c|}{24.45}      & 21.57  & 21.49      \\
WISE-FT \cite{wortsman2022robust}
& 27.08      & 27.85      & 16.53      & \multicolumn{1}{c|}{29.17}      & 25.16    & 24.67
& 23.43      & 24.00      & 16.01      & \multicolumn{1}{c|}{28.02}      & 22.87    & 24.63    \\
L2P \cite{wang2022learning}
& 70.73      & 72.96      & 32.25      & \multicolumn{1}{c|}{39.54}      & 53.87   & 11.25    
& 72.19      & 62.37      & 37.03      & \multicolumn{1}{c|}{43.38}      & 53.74   & 11.95     \\
S-liPrompts \cite{wang2022s}
& 67.93      & 75.45      & 40.88      & \multicolumn{1}{c|}{43.25}      & 56.88   & 12.02
& 67.81      & 73.42      & 32.17      & \multicolumn{1}{c|}{38.62}      & 53.01   & 11.29     \\
AttriCLIP \cite{wang2023attriclip}
& 68.62      & 72.69      & 42.30      & \multicolumn{1}{c|}{42.24}      & 56.46   & 10.43     
& 67.89      & 69.91      & 38.26      & \multicolumn{1}{c|}{44.05}      & 55.03   &  12.62    \\
PGP \cite{qiao2024PGP}
& 72.51      & 76.51      & 37.42     & \multicolumn{1}{c|}{43.87}      & 57.58   & 10.11     
& 72.15      & 66.24      & 37.22     & \multicolumn{1}{c|}{42.72}      & 54.58   & 11.31     \\
\rowcolor[gray]{0.9}COMM (Ours)
& \textbf{81.95}      & \textbf{85.17}      & \textbf{50.39}      & \multicolumn{1}{c|}{\textbf{61.01}}      & \textbf{69.63}      & 7.81    
& \textbf{80.35}      & \textbf{84.99}      & \textbf{50.30}      & \multicolumn{1}{c|}{\textbf{60.94}}      & \textbf{69.14}      & 7.94  \\\hline
\end{tabular}%
}
\caption{Results for the second sequence.
AIA and F are measured after training on all tasks.
Overall AIA and Overall F represent the average AIA and average F across all modalities, respectively.}
\label{tab: Scenario - S2}
\end{table*}
\setlength{\tabcolsep}{6pt}
\vspace{-2mm}
\renewcommand{\arraystretch}{1.0}

\begin{figure*}[] 
     \centering \includegraphics[width=0.98\textwidth]{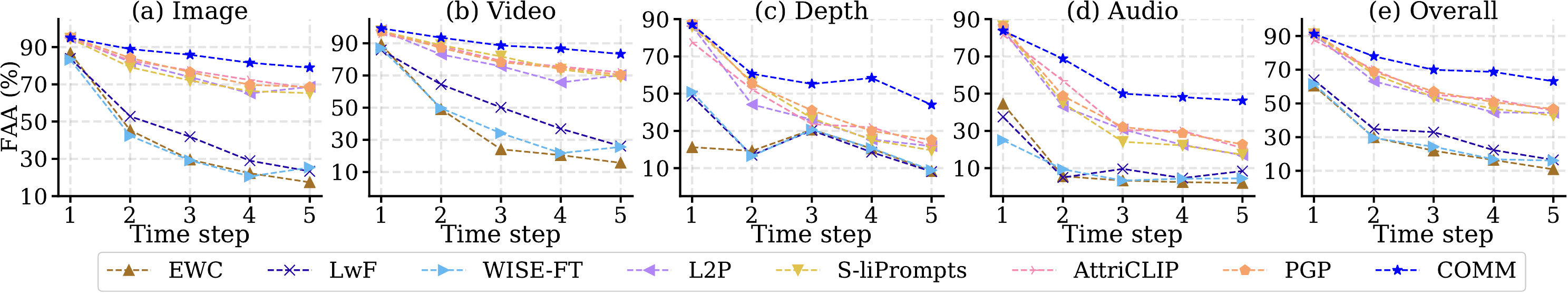}
    \caption{Results (measured in FAA) of multimodal continual learning for the third sequence.}
    \vspace{-2mm}
    \label{fig: Sequence S3} 
\end{figure*}

\subsection{Main Results}
\label{section: Main results - S1 and S2}
\noindent\textbf{Sequence 1.}
We evaluated our method and the compared methods on the first sequence for multimodal continual learning.
Table \ref{tab: Scenario - S1} summarizes the results of the modality-specific (top) and modality-agnostic (bottom) class-incremental learning scenarios.
Overall, we observe that the results of the modality-agnostic scenario show lower performance compared to the modality-specific scenario due to the challenge in identifying the modality.
The regularization methods, EWC, LwF, and WISE-FT, perform poorly in both scenarios compared to the prompt-based continual learning methods, L2P, S-liPrompts, AttriCLIP, PGP, and COMM.
This indicates that directly updating the network parameters in the regularization methods to accommodate knowledge from multiple modalities is prone to forgetting previously learned tasks due to contaminated mixed knowledge.
Even though the existing prompt-based methods outperform other continual learning approaches, they do not perform well for the depth and audio modalities. 
This is mainly due to the absence of pre-training on non-visual data, which leads to greater forgetting; the proposed method alleviates this by effectively weighting prompts and suppressing cross-modal interference.
Notably, these methods exhibit lower AIA in the modality-agnostic scenario due to the challenge of selecting prompts from the shared pool, which hinders learning new tasks.
The proposed method, aiming to alleviate sensitivity to prompt selection and biased projections, outperforms the best competitor, S-liPrompts, with an overall FAA gap of 16.85\% and 13.56\% in the respective scenarios.
In terms of the performance gap between the scenarios, the proposed method shows only a marginal performance gap of 0.63\% in overall AIA, in contrast to the larger gaps of 4.41\% and 6.93\% observed with S-liPrompts and AttriCLIP, respectively.
Additional results for different class orders can be seen in the supplementary material.

We also present detailed results of the modality-agnostic scenario (measured in FAA) of the compared methods in Figure \ref{fig: S1 modality-agnostic}.
The conventional methods that directly update the network parameters exhibit high fluctuations.
The performance of each modality slightly increases when tasks with the specific modality are learned (indicated by vertical lines). 
However, learning tasks from other modalities (indicated by non-vertical lines) results in a significant performance decline due to overwritten knowledge.
The prompt-based continual learning methods, L2P, S-liPrompts, AttriCLIP, and PGP, also show unsatisfying performance in video, depth, and audio modalities.
In contrast, the proposed method exhibits stable performance across different modalities, surpassing other approaches, particularly in the audio and depth modalities.

\renewcommand{\arraystretch}{0.8}
\begin{table*}[t!]
\centering
\scriptsize
\setlength\doublerulesep{.5pt}
\resizebox{0.9\textwidth}{!}{%
\begin{tabular}{ccc||ccccc|ccccc}
\hline
\multicolumn{3}{c||}{Method} & \multicolumn{5}{c|}{First Sequence (Random)} & \multicolumn{5}{c}{Second Sequence (Shift)} \\ \cline{1-13} 
{\textit{Cross}} & \textit{Self} & \textit{Re-align}
& Image  & Video  & Depth   & \multicolumn{1}{c|}{Audio}  & Overall 
& Image  & Video  & Depth   & \multicolumn{1}{c|}{Audio}  & Overall \\ \hline
\checkmark & - & -  
& 62.81       & 80.42       & 46.44      & \multicolumn{1}{l|}{49.84}      
& 59.88       
& 59.77       & 70.14       & 23.35      & \multicolumn{1}{l|}{43.76}      
& 49.26 \\ 
- & \checkmark & -  
& 73.95       & 75.14       & 48.92      & \multicolumn{1}{l|}{59.35}      
& 64.34       
& 65.52       & 65.94       & 30.75      & \multicolumn{1}{l|}{56.24}      
& 54.61  \\ 
- & - & \checkmark 
& 76.10       & 81.50       & 49.32      & \multicolumn{1}{l|}{53.65}      
& 65.14       
& 67.81       & 76.88       & 31.92      & \multicolumn{1}{l|}{51.11}      
& 56.93  \\ 
\checkmark & \checkmark & -  
& 71.91       & 73.21       & 48.51      & \multicolumn{1}{l|}{59.53}      
& 63.29
& 64.08       & 65.78       & 41.76      & \multicolumn{1}{l|}{54.46}      
& 56.53  \\ 
\checkmark & - & \checkmark
& 76.40       & 81.47       & 58.01      & \multicolumn{1}{l|}{54.99}      
& 67.72        
& 66.73       & 76.63       & 25.65      & \multicolumn{1}{l|}{49.77}      
& 54.70  \\ 
-  & \checkmark & \checkmark
& 83.05      & 88.62     & 47.55      & \multicolumn{1}{l|}{60.59}      
& 69.95   
& 78.17       & 81.51       & 29.92      & \multicolumn{1}{l|}{60.83}      
& 62.61 \\ 
\rowcolor[gray]{0.9} \checkmark & \checkmark & \checkmark 
& \textbf{83.89}      & \textbf{90.83}       & \textbf{62.78}      & \multicolumn{1}{l|}{\textbf{61.85}}      
& \textbf{74.84} 
& \textbf{80.35}       & \textbf{84.99}       & \textbf{50.30}      & \multicolumn{1}{l|}{\textbf{60.94}}      
& \textbf{69.14}  \\ 
\hline
\end{tabular}%
}
\caption{Results of the ablation study on the usage of the components in the proposed method. 
}
\label{tab: ablation study components}
\end{table*}
\renewcommand{\arraystretch}{1.0}

\noindent{\textbf{Sequence 2.}}
To further evaluate the consistency of our approach, we compared COMM with other continual learning methods in another sequence, where each modality is trained over multiple times before shifting to another (in order of image, video, depth, and audio).
Table \ref{tab: Scenario - S2} reports the experimental results for the sequence, demonstrating a similar performance trend to that observed in Table \ref{tab: Scenario - S1}.
However, conventional continual learning methods show a significant performance degradation for modalities learned early.
Specifically, the image and video modalities experience about a 20\% drop in AIA compared to the first sequence.
Similarly, prompt-based methods show performance declines for previously learned modalities; however, they achieve enhanced performance for the most recently acquired modality, audio.
The results indicate that learning different modalities increases interference between modalities, leading to more forgetting in both continual learning methods.
COMM outperforms PGP and AttriCLIP in each scenario, demonstrating significant AIA improvements in audio with gaps of 17.14\% and 16.89\%, respectively.
Additionally, it achieves overall AIA improvements of 12.05\% (with an overall F decrease of 2.30\%) and 14.11\% (with an overall F decrease of 4.68\%) compared to PGP and AttriCLIP.
We also provide additional results on sequences with varying task orders in the supplementary material.

\noindent\textbf{Sequence 3.}
We report the results for multiple modalities learned simultaneously at every step in a modality-specific scenario.
Figure \ref{fig: Sequence S3} shows the results for the sequence. 
The results reveal a clear difference between methods using prompts (L2P, S-liPrompts, AttriCLIP, and PGP) and those updating model parameters (EWC, LwF, and WISE-FT) across all time steps.
This indicates that directly updating model parameters for multiple modalities can lead to severe conflicts, resulting in poor performance.
The proposed approach consistently demonstrates higher performance across all modalities than the competitors due to reduced interference between modalities.

\subsection{More Results}
\noindent{\textbf{Ablation study on the components}.}
We conducted an ablation study of the proposed method, including its components: cross-modality integration ({\textit{Cross}}), self-regularization (\textit{Self}), and projection head re-alignment (\textit{Re-align}).
Note that the method without \textit{Cross} or \textit{Self} requires identifying the modality identity or task to select prompts, respectively. 
Therefore, we employed a prompt selection mechanism in \cite{wang2022s}.
We compared these methods to the first two sequences in the modality-agnostic scenario.
Table \ref{tab: ablation study components} reports the results of the ablation study with AIA measured after training all tasks.
Overall, the proposed method using an individual component results in unsatisfactory performance on both task sequences compared to employing them together.
The method using \textit{Cross} identifies relevant modalities but struggles to select accurate task-specific prompts.
When comparing the method utilizing all components against that excluding \textit{Cross}, it becomes evident that selecting suitable modality prompts significantly enhances performance, with gaps of 15.23\% and 20.38\% in the depth modality for the first and second sequences, respectively.
The method that employs both \textit{Self} and \textit{Re-align} exhibits a strong synergistic effect because \textit{Self} preserves previous features by regularizing prompts, and these preserved features closely resemble those used by \textit{Re-align}.
We provide visualizations demonstrating the maintainability of \textit{Self} in the embedding space in the supplementary material.

\begin{figure}[t] 
     \centering \includegraphics[width=\columnwidth]{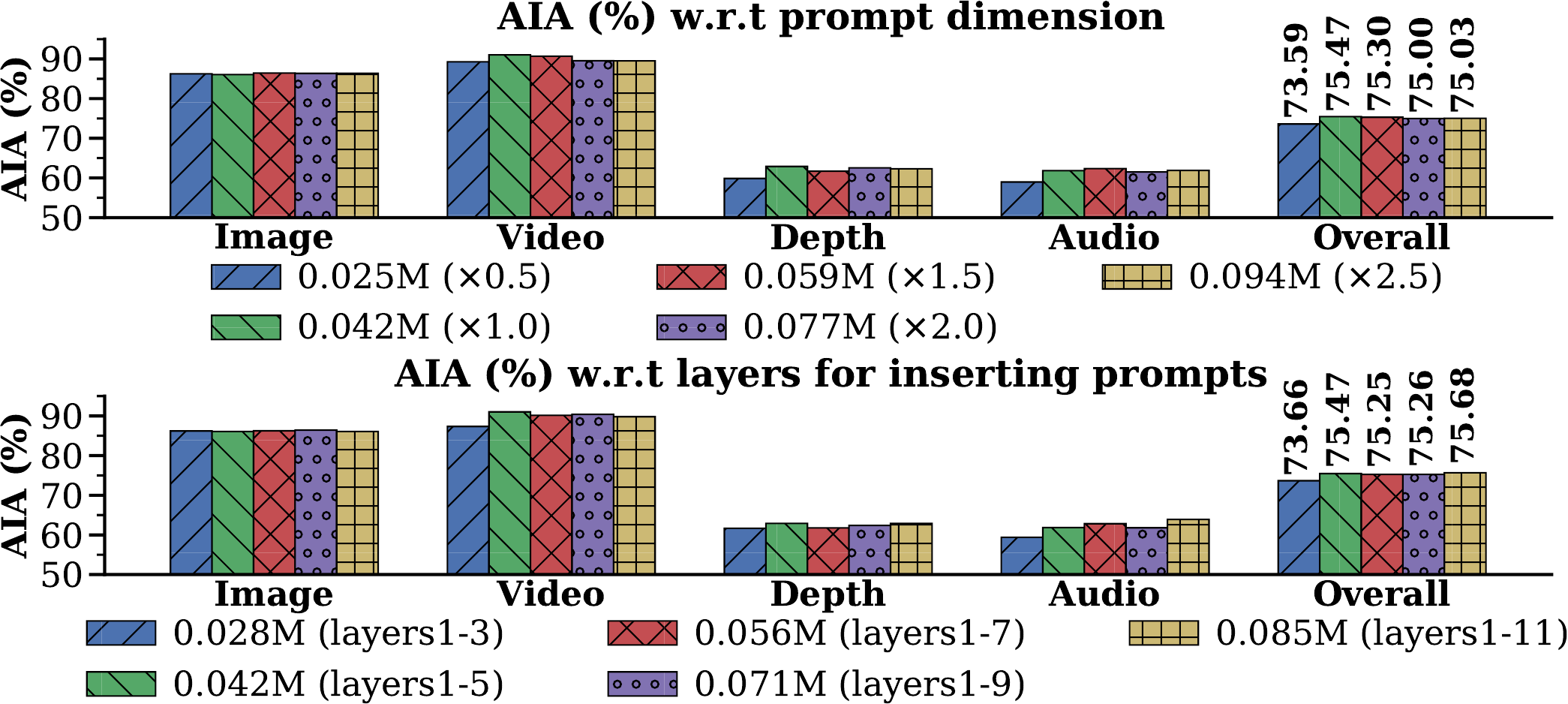}
    \caption{Results for the proposed method with varying learnable parameters by adjusting the prompt dimension (top) and the layers where prompts are inserted (bottom).}
    \label{fig: param-acc} 
\end{figure}

\begin{figure}[t] 
     \centering \includegraphics[width=\columnwidth]{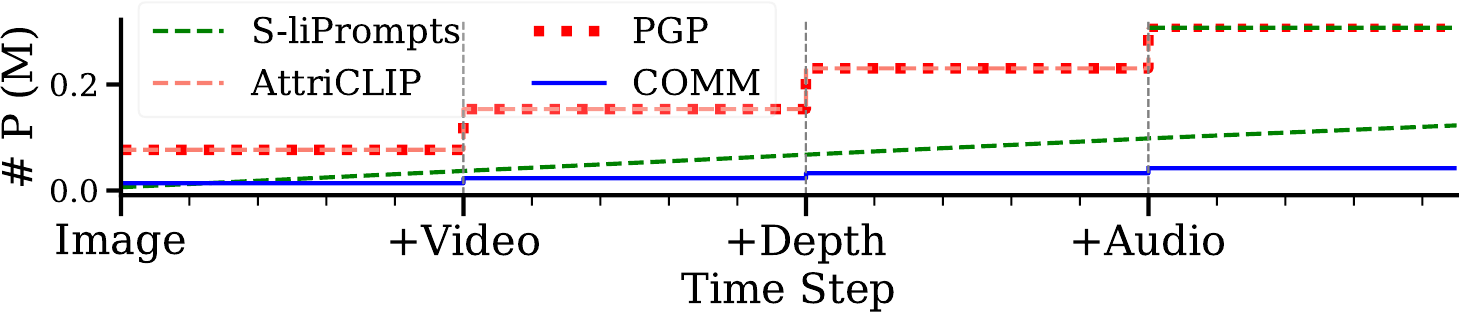}
    \caption{The number of parameters with respect to the modality.}
    \label{fig: param growth} 
\end{figure}

\noindent{\textbf{Study of scalability. }}
In this study, we evaluated the scalability of the proposed method by examining the impact of varying numbers of learnable parameters.
To adjust the number of learnable parameters, we increased the dimension of prompts and the layers of the pre-trained models for inserting prompts, following the methods in the previous studies \cite{khattak2023maple, zhou2022conditional}. 
Figure \ref{fig: param-acc} indicates that COMM maintains consistent overall AIA performance across different prompt dimensions and numbers of layers for prompt insertion.
Furthermore, we also compared the parameter growth of COMM with other prompt-based methods in Figure \ref{fig: param growth}. 
Overall, prompt-based methods require fewer parameters than updating the entire model \cite{kirkpatrick2017overcoming, li2016learning}, which involves approximately 124 million parameters.
Specifically, L2P and AttriCLIP leverage a prompt pool for each modality, resulting in a step-wise increase in parameters as new modalities are introduced. 
S-liPrompts uses task-specific prompts, leading to a monotonic parameter increase.
The proposed method, COMM, consolidates prompts inter- and intra-modalities, enhancing scalability when integrating additional modalities.
Additional results on prompting designs, prompt accumulation strategies, and unimodal continual learning are included in the supplementary material.

\section{Conclusion}
In this work, we have introduced the first multimodal continual learning framework that learns tasks of different modalities.
Learning multiple tasks complicates the retention of prior knowledge as accumulating mixed information over time interferes with the learning of individual modalities.
To address this challenge, we have proposed a strategy that aggregates knowledge in each modality over time through self-regularization and integrates it across relevant modalities.
This approach selectively emphasizes contributions from relevant modalities while minimizing the influence of unrelated ones, gradually assimilating new information and preserving existing knowledge.
Extensive experimental results, including modality-specific and modality-agnostic scenarios, demonstrated that ours significantly outperforms other continual learning methods with a notable performance gap.

\bibliography{aaai2026}

\clearpage
\setcounter{table}{0}
\setcounter{figure}{0}
\renewcommand\thetable{\Alph{table}}
\renewcommand\thefigure{\Alph{figure}}
\section{Supplementary material}

\subsection{Additional Implementation Details}
\noindent{\textbf{Additional details of datasets.}}
The datasets used for the modalities are ImageNet-100 \cite{deng2009imagenet}, UCF-101 \cite{soomro2012ucf101}, ESC-50 \cite{piczak2015esc}, and SUN-RGBD \cite{song2015sun}, each accompanied by corresponding text data.
ImageNet-100 contains 100 classes selected from the original ImageNet dataset \cite{deng2009imagenet}, widely used in continual learning for image classification \cite{douillard2022dytox, yan2021dynamically}.
UCF-101 contains 101 action categories for video action recognition \cite{soomro2012ucf101}.
ESC-50 is a fine-grained collection of environmental audio recordings of 50 classes for sound classification.
For the depth modality, we performed on the SUN-RGBD dataset for scene classification.

\vspace{1mm}
\noindent{\textbf{Preprocessing multiple modalities.}}
Images from ImageNet-100 were resized to a spatial dimension of 224$\times$224 pixels.
We applied center cropping for augmentation. 
For SUN-RGBD, we converted the depth maps into disparity maps following \cite{girdhar2022omnivore} and normalized the disparity map between 0 and 1.
For the video dataset, we randomly selected three frames from every video clip of two seconds.
The height and width of each video frame were resized to $224\times224$ pixels, resulting in dimensions of $6\times3\times224\times224$. 
We augmented videos with center cropping and horizontal flipping following \cite{villa2022vclimb}. 
For the audio dataset, we selected three audio clips from each audio sample. 
We processed each audio clip by sampling it at 16kHz, followed by extracting a mel-spectrogram with 224 frequency bins using a 25ms Hamming window every 10ms \cite{girdhar2023imagebind}, and resized the spatial dimension to $224\times224$, resulting in dimensions of $3\times3\times224\times224$. 
We applied the same processing and augmentation for the modalities across all comparison methods.
The preprocessed non-text modalities were tokenized using the tokenizer of the pre-trained vision encoder \cite{radford2021learning}, following previous practices \cite{girdhar2023imagebind, zhu2023languagebind}.

\renewcommand{\arraystretch}{1.0}
\begin{table*}[t!]
\centering
\setlength\doublerulesep{.5pt}
\resizebox{\textwidth}{!}{%
\begin{tabular}{l||cccccc|cccccc}
\hline
\multirow{3}{*}{Method} & \multicolumn{6}{c|}{Reverse Order of First Sequence (Random)} & \multicolumn{6}{c}{Reverse Order of Second Sequence (Shift)} \\ \cline{2-13} 
& Audio ($\uparrow$) & Depth ($\uparrow$) & Video ($\uparrow$) & \multicolumn{1}{c|}{Image ($\uparrow$)} & \begin{tabular}[c]{@{}c@{}}Overall\\ AIA ($\uparrow$)\end{tabular} & \begin{tabular}[c]{@{}c@{}}Overall\\ F ($\downarrow$)\end{tabular}
& Audio ($\uparrow$) & Depth ($\uparrow$) & Video ($\uparrow$) & \multicolumn{1}{c|}{Image ($\uparrow$)} & \begin{tabular}[c]{@{}c@{}}Overall\\ AIA ($\uparrow$)\end{tabular} & \begin{tabular}[c]{@{}c@{}}Overall\\ F ($\downarrow$)\end{tabular} \\ \hline
FT 
& 27.18      & 26.65      & 9.13      & \multicolumn{1}{c|}{6.33}      & 17.32 & 8.75
& 12.24      & 16.89      & 20.13      & \multicolumn{1}{c|}{38.98}      & 22.06 & 17.93       \\
EWC \cite{kirkpatrick2017overcoming}
& 33.06      & 28.76      & 9.34      & \multicolumn{1}{c|}{2.22}      & 18.35  & 9.84   
& 12.44      & 16.56     & 23.78      & \multicolumn{1}{c|}{39.42}      & 23.05   & 18.25     \\
LwF \cite{li2016learning}
& 26.12      & 22.26      & 13.88      & \multicolumn{1}{c|}{8.06}      & 17.58   & 11.35
& 12.90      & 17.45       & 24.67      & \multicolumn{1}{c|}{39.91}      & 23.73  & 19.16      \\
WISE-FT \cite{wortsman2022robust}
& 28.48      & 28.13      & 8.71      & \multicolumn{1}{c|}{4.57}      & 17.47 & 9.63
& 14.16      & 17.42      & 24.67     & \multicolumn{1}{c|}{38.57}      & 23.71 & 16.81       \\
L2P \cite{wang2022learning}
& 42.43      & 44.01      & 69.35      & \multicolumn{1}{c|}{62.80}      & 54.65  & 19.29  
& 24.98      & 31.01      & 66.75      & \multicolumn{1}{c|}{77.64}      & 50.10  & 6.68      \\
S-liPrompts \cite{wang2022s}
&  44.03      & 45.94      & 74.03      & \multicolumn{1}{c|}{61.14}      & 56.29  & 20.01
&  24.85      & 31.18      & 71.65      & \multicolumn{1}{c|}{75.10}      & 50.70  & 12.04   \\
AttriCLIP \cite{wang2023attriclip}
& 44.88      & 42.62      & 68.31      & \multicolumn{1}{c|}{58.90}      & 53.68  & 20.89
& 26.61      & 28.96      & 68.92      & \multicolumn{1}{c|}{72.94}      & 49.36  &  7.59     \\
PGP \cite{qiao2024PGP}
& 41.20      & 49.40      & 79.63      & \multicolumn{1}{c|}{79.30}      & 62.39  & 17.05      
& 26.08      & 33.56      & 62.13      & \multicolumn{1}{c|}{79.82}      & 50.40  & 11.37 \\
\rowcolor[gray]{0.9}COMM (Ours)
& \textbf{63.08}      & \textbf{61.22}      & \textbf{89.14}      & \multicolumn{1}{c|}{\textbf{87.03}}      & \textbf{75.12}      &  10.47  
& \textbf{53.02}      & \textbf{53.42}      & \textbf{84.62}      & \multicolumn{1}{c|}{\textbf{85.88}}      & \textbf{69.23}      &  4.96 \\\hline
\end{tabular}%
}
\caption{{Results for the reverse task order of the first (left) and second (right) sequences in the modality-agnostic class-incremental learning scenario.}
The AIA and F is measured after training on all tasks.
Overall AIA and Overall F represent the average AIA and average F across all modalities.}
\label{tab: reverse S1 S2}
\end{table*}
\renewcommand{\arraystretch}{1.0}

\renewcommand{\arraystretch}{1.0}
\begin{table*}[t!]
\centering
\begin{tabular}{p{0.47\textwidth}|p{0.47\textwidth}} 
\hline
\textbf{Image} & \textbf{Video} \\ \hline
great white shark, drake, electric ray, mud turtle, sea lion, wallaby, conch, American coot, oystercatcher, barn spider, hognose snake, great grey owl, coucal, diamondback, prairie chicken, sea slug, black swan, black and gold garden spider, spiny lobster, leatherback turtle, black grouse, bittern, common newt, white stork, tiger shark, kite (bird of prey), tick, pelican, garden spider, green mamba, horned viper, sea anemone, snail, crayfish, vine snake, sulphur-crested cockatoo, hen, redshank, jellyfish, macaw, chambered nautilus, wolf spider, water ouzel, stingray, Dungeness crab, hammerhead, hermit crab, scorpion, crane, goldfinch, bustard, lorikeet, thunder snake, goldfish, tarantula, cock, agama, goose, ptarmigan, sea snake, flamingo, green snake, toucan, axolotl, chickadee, peacock, night snake, hornbill, tailed frog, common iguana, boa constrictor, red-backed sandpiper, Komodo dragon, spoonbill, black widow, bulbul, whiptail, rock crab, American alligator, flatworm, banded gecko, sidewinder, chiton, spotted salamander, magpie, bee eater, king snake, terrapin, hummingbird, wombat, garter snake, green lizard, tench, albatross, nematode, harvestman, bald eagle, indigo bunting, loggerhead, limpkin.
& surfing, salsa spin, playing tabla, pole vault, yo yo, mixing, apply eye makeup, volleyball spiking, long jump, field hockey penalty, rafting, hammering, push ups, handstand walking, blowing candles, golf swing, playing guitar, band marching, brushing teeth, playing flute, front crawl, archery, bench press, punch, horse riding, cutting in kitchen, billiards, cricket bowling, diving, haircut, boxing speed bag, horse race, still rings, high jump, biking, ice dancing, skijet, writing on board, hula hoop, throw discus, parallel bars, playing daf, wall pushups, swing, hammer throw, balance beam, typing, breast stroke, tennis swing, handstand pushups, sky diving, knitting, rope climbing, skiing, basketball dunk, baby crawling, rowing, nunchucks, table tennis shot, trampoline jumping, bowling, basketball, drumming, skate boarding, lunges, baseball pitch, kayaking, cliff diving, frisbee catch, rock climbing indoor, soccer juggling, shot put, playing piano, javelin throw, military parade, boxing punching bag, sumo wrestling, soccer penalty, pommel horse, walking with dog, clean and jerk, apply lipstick, playing violin, jumping jack, fencing, pull ups, juggling balls, body weight squats, pizza tossing, head massage, playing dhol, shaving beard, mopping floor, floor gymnastics, blow dry hair, jump rope, playing cello, playing sitar, tai chi, uneven bars, cricket shot. \\ \hline
\textbf{Depth} & \textbf{Audio} \\ \hline
dancing room, computer room, coffee room, 
lobby, office dining, idk, storage room, bedroom, 
lecture theatre, library, corridor, reception room, 
playroom, classroom, cafeteria, gym, study space, 
indoor balcony, office, basement, living room, rest space, 
dinette, dining room, lab, furniture store, hotel room, 
printer room, bathroom, stairs, kitchen, reception, study, 
dining area, bookstore, exhibition, discussion area, home, 
mail room, laundromat, home office, recreation room, 
office kitchen, conference room. & 
airplane, wind, clock tick, toilet flush, sneezing, 
rooster, sheep, pig, door wood creaks, hand saw, 
washing machine, crying baby, train, clock alarm, frog, 
rain, siren, crickets, snoring, helicopter, keyboard typing, 
cow, laughing, church bells, coughing, can opening, 
water drops, vacuum cleaner, insects, cat, door wood knock, 
thunderstorm, chainsaw, crackling fire, car horn, brushing teeth, 
footsteps, clapping, pouring water, mouse click, fireworks, 
breathing, sea waves, glass breaking, hen, chirping birds, 
crow, dog, drinking sipping, engine. \\ \hline
\end{tabular}
\caption{Class orders (represented by their names) for each modality dataset.}
\label{tab:class order}
\end{table*}
\renewcommand{\arraystretch}{1.0}

\vspace{1mm}
\noindent{\textbf{Implementation details.}}
We inserted prompts up to the fifth layer of each encoder, following \cite{khattak2023maple}.
Additionally, due to the high dimension of the projection layer $\alpha^{t}_{m}$, we employed a low-rank decomposition technique \cite{hu2021lora} to split it into two smaller matrices with a rank of 20.
To train modalities involving temporal information (video and audio), we duplicated the prompts for the temporal dimensions and concatenated them with the spatial features of each temporal dimension \cite{rasheed2023fine}.
We report the average results from three independent runs for all experiments using NVIDIA RTX A5000 GPUs.

The modality features at time $t$, $V_{\hat{P}^{t}_{m}}(x^{t}_{m,i})$, were average pooled along the temporal dimension before being fed into the projection head.
To train the layer $g_{\beta^{t}}$, we computed and stored the mean and covariance of the features for each modality at the beginning of every time step.
Specifically, for the $i$-th sample of modality $m$, $x^{t}_{m,i}$, the feature was calculated as $V(x^{t}_{m,i})$. 
For modality $m$, if the mean and covariance had already been stored from previous tasks, the mean was updated by combining the previous and new means based on their sample sizes, and the covariance was recomputed using the updated mean and new statistics.
Finally, we trained the function $g_{\beta^{t}}$ using the sampled features to capture the relevance between modalities.

\renewcommand{\arraystretch}{1.0} 
\begin{table}[t!]
\centering
\Large
\resizebox{\columnwidth}{!}{%
\begin{tabular}{c|c|c|c|c|c|>{\columncolor[gray]{0.93}}c}
\hline
\multicolumn{2}{c|}{Method}                    
& L2P  & S-liPrompts  & AttriCLIP  & PGP  & COMM               \\ \hline
\hline
\multirow{3}{*}{\rotatebox{90}{Image}} 
& AIA ($\uparrow$)
& 79.3 $\pm$ 2.1 
& 76.4 $\pm$ 0.6 
& 77.9 $\pm$ 2.4 
& 81.0 $\pm$ 1.8
& \textbf{85.9 $\pm$ 0.3} \\ 
& FAA  ($\uparrow$)                  
& 70.9 $\pm$ 2.4 
& 67.7 $\pm$ 1.8 
& 66.6 $\pm$ 3.0 
& 71.4 $\pm$ 2.6
& \textbf{79.6 $\pm$ 0.4} \\ 
& F    ($\downarrow$)
& 8.8  $\pm$ 0.5 
& 9.1  $\pm$ 1.4 
& 11.9 $\pm$ 0.7 
& 9.3  $\pm$ 1.1
& \textbf{6.3 $\pm$ 1.0} \\ \hline
\multirow{3}{*}{\rotatebox{90}{Video}} 
& AIA ($\uparrow$)
& 80.6 $\pm$ 2.1 
& 78.6 $\pm$ 3.9 
& 80.4 $\pm$ 2.3 
& 80.6 $\pm$ 2.5
& \textbf{90.1 $\pm$ 0.7} \\ 
& FAA  ($\uparrow$)              
& 69.8 $\pm$ 2.1 
& 65.2 $\pm$ 3.0 
& 72.0 $\pm$ 1.2 
& 70.6 $\pm$ 0.6
& \textbf{82.9 $\pm$ 0.6} \\ 
& F    ($\downarrow$)                  
& 11.5 $\pm$ 2.3 
& 14.2 $\pm$ 1.3 
& 8.9 $\pm$ 1.2 
& 12.1 $\pm$ 1.8
& \textbf{7.7 $\pm$ 2.8} \\ \hline
\multirow{3}{*}{\rotatebox{90}{Depth}}
& AIA ($\uparrow$)
& 40.3 $\pm$ 3.8 
& 44.5 $\pm$ 2.6 
& 44.4 $\pm$ 4.8 
& 46.2 $\pm$ 2.2
& \textbf{61.8 $\pm$ 2.7} \\ 
& FAA  ($\uparrow$)              
& 19.7 $\pm$ 4.4 
& 23.5 $\pm$ 1.8 
& 28.2 $\pm$ 4.8 
& 29.1 $\pm$ 3.1
& \textbf{44.9 $\pm$ 1.0} \\ 
& F    ($\downarrow$)
& 22.9  $\pm$ 1.4 
& 24.5  $\pm$ 3.9 
& 17.1  $\pm$ 1.1 
& 18.1  $\pm$ 5.2
& \textbf{16.8  $\pm$ 3.8} \\ \hline
\multirow{3}{*}{\rotatebox{90}{Audio}}
& AIA ($\uparrow$)
& 46.7 $\pm$ 1.8
& 47.9 $\pm$ 4.5 
& 46.4 $\pm$ 2.1 
& 47.0 $\pm$ 0.9
& \textbf{65.7 $\pm$ 2.9} \\
& FAA  ($\uparrow$)                  
& 19.5 $\pm$ 1.9 
& 18.8 $\pm$ 2.9 
& 20.5 $\pm$ 1.3 
& 22.3 $\pm$ 3.0
& \textbf{47.9 $\pm$ 3.4} \\
& F    ($\downarrow$)                  
& 28.9 $\pm$ 0.5 
& 30.9 $\pm$ 6.1 
& 27.5 $\pm$ 2.7 
& 28.0 $\pm$ 2.1 
& \textbf{18.9 $\pm$ 1.8} \\ \hline
\multirow{3}{*}{\rotatebox{90}{Overall}}
& AIA ($\uparrow$)
& 61.7 $\pm$ 2.4
& 61.9 $\pm$ 2.9
& 62.3 $\pm$ 2.9
& 63.7 $\pm$ 1.8
& \textbf{75.9 $\pm$ 1.6} \\
& FAA  ($\uparrow$)                  
& 45.0 $\pm$ 2.7
& 43.8 $\pm$ 2.4
& 46.8 $\pm$ 2.5
& 48.3 $\pm$ 2.4
& \textbf{63.8 $\pm$ 1.3} \\
& F    ($\downarrow$)                  
& 18.0 $\pm$ 1.1  
& 19.7 $\pm$ 3.2 
& 16.3 $\pm$ 1.4
& 16.8 $\pm$ 2.5
& \textbf{12.4 $\pm$ 1.7} \\ \hline
\end{tabular}%
}
\caption{Results for three different class orders (mean$\pm$std).}
\label{tab:rebuttal}
\renewcommand{\arraystretch}{1} 
\end{table}

\subsection{Additional Results}
\noindent{\textbf{Results on different task orders.}}
For the experiment conducted in the Table \ref{tab: Scenario - S1} in the main paper, the sequence consists of 20 tasks, where image, video, depth, and audio modalities were learned repetitively.
For the sequence in Table \ref{tab: Scenario - S2}, modality was trained for five time steps before shifting to next modality in the order of image, video, depth, and audio.
To investigate the sensitivity of the compared methods with respect to different task orders in continual learning, we conducted an additional experiments using the reverse order of the tasks used in Tables \ref{tab: Scenario - S1} and \ref{tab: Scenario - S2}.

The results are reported in Table \ref{tab: reverse S1 S2}.  
The conventional continual learning methods, EWC, LwF, and WISE-FT, are highly sensitive to the order of tasks for both sequences.
In contrast, the prompt-based methods, L2P, S-liPrompt, AttriCLIP, PGP, and COMM, show less sensitivity to the order of tasks compared to the conventional methods.
The proposed method, COMM, demonstrates robustness to changes in task order, consistently outperforms other continual learning methods.
Specifically, the proposed method shows a marginal overall AIA gap of 0.09\% between the different task orders in the second sequence.
However, other prompt-based continual learning methods, L2P, S-liPrompts, AttriCLIP, and PGP, show larger performance gaps of 3.64\%, 2.31\%, 5.67\%, and 4.18\%, respectively.

\begin{figure*}[t] 
     \centering \includegraphics[width=\textwidth]{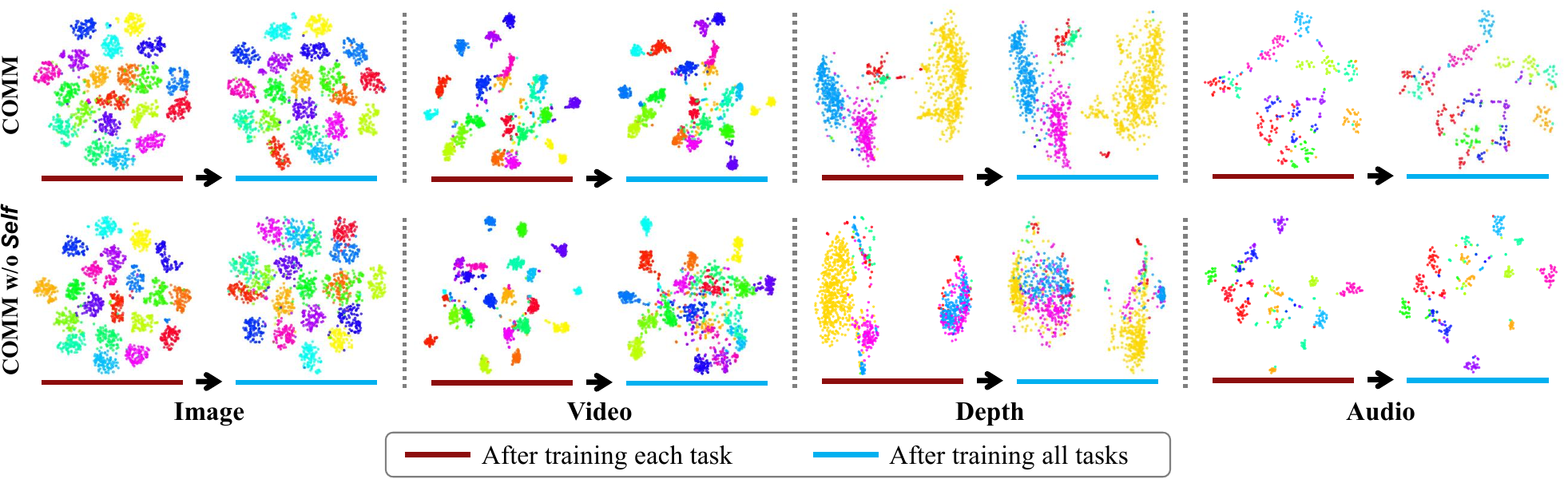}
    \caption{t-SNE visualization on the modality embeddings extracted from the proposed method, COMM, with and without including self-regularization (\textit{Self}) of prompts. 
    Each color corresponds to an individual class. 
    }
    \label{fig: t-sne}
\end{figure*}

\vspace{1mm}
\noindent{\textbf{Results on different class orders.}}
We followed the class order provided by \cite{villa2022vclimb} for the video dataset.
Similar to the practices in other continual learning benchmarks \cite{villa2022vclimb, rebuffi2017icarl, wang2023attriclip}, we randomly selected the class order for the image \cite{deng2009imagenet}, depth \cite{song2015sun}, and audio \cite{piczak2015esc} datasets.
Based on the determined class orders, the image and audio datasets were divided into five subsets consisting of 20 and 10 classes, respectively. 
The video dataset was divided into five subsets, with the first subset comprising 21 classes and the other four subsets containing 20 classes each.
Similarly, the depth dataset was divided into five subsets, with the first four subsets comprising eight classes each and the remaining subset containing 12 classes.
The class orders for the image, video, depth, and audio data used in the main paper are listed in Table \ref{tab:class order}.

To validate the robustness of the proposed method under different class orders, we conducted additional experiments using three randomly shuffled class orders in the modality-specific scenario for the first sequence.
Since regularization-based continual learning approaches exhibit unsatisfactory performance, we omit them from the comparison.
Table \ref{tab:rebuttal} shows AIA, FAA, and Forgetting (F) results after training on all tasks. 
COMM consistently outperforms all other methods across all modalities in terms of AIA and FAA while exhibiting the lowest forgetting. 
Specifically, COMM outperforms the runner-up method, PGP, with performance margins of 12.2\%, 15.5\%, and 4.4\% in AIA, FAA, and F, respectively.
These results demonstrate the proposed COMM's robustness to class order variations.

\begin{table*}[t!]
\centering
\setlength\doublerulesep{0.5pt}
\resizebox{0.9\textwidth}{!}{
\begin{tabular}{l|c|c|c||cccc|c}
\hline
\rowcolor{gray!15}
Variant & Prompt Insertion & Intra-modal & Inter-modal & Image & Video & Depth & Audio & Overall \\
\hline
COMM-VL (Ours)      & Both encoders  & Self-Regulation     & Learned relevance     & \textbf{86.07} & \textbf{91.03} & \textbf{62.90} & \textbf{61.86} & \textbf{75.47} \\
\hline
COMM-L       & Language encoder      & Self-Regulation     & Learned relevance     & 85.30 & 83.89 & 61.56 & 60.43 & 72.80 \\
COMM-V       & Modality encoder  & Self-Regulation     & Learned relevance     & 85.76 & 90.50 & 59.22 & 61.35 & 74.21 \\
COMM-Sum     & Both encoders  & Sum          & Learned relevance     & 78.00 & 77.70 & 51.10 & 52.90 & 65.00 \\
COMM-Avg     & Both encoders  & Avg          & Learned relevance     & 76.90 & 76.60 & 51.70 & 54.40 & 64.90 \\
COMM-NP & Both encoders & Self-Regulation     & Mahalanobis & 85.10 & 84.70 & 57.80 & 59.80 & 71.80 \\
\hline
\end{tabular}
}
\caption{Ablation study on prompt configuration. We compare variations in (a) insertion location, (b) intra-modality prompt accumulation strategy, and (c) inter-modal prompt integration method.}
\label{tab:ablation_prompt_config}
\end{table*}

\newpage
\noindent{\textbf{Impact of self-regularization.}}
The proposed knowledge aggregation across time within a modality consolidates old and new knowledge by self-regulating the previously aggregated prompts and the current ones.
This approach eliminates the need to store previous individual prompts $\{P^{1}_{m},\dots,P^{t-1}_{m}\}$ and $\{Q^{1}_{m},\dots,Q^{t-1}_{m}\}$, which provides computational efficiency compared to other prompt-based continual learning methods \cite{wang2022s, zhou2022learning, smith2023coda}.
This choice of implementation is particularly important in multimodal continual learning, where new tasks can introduce novel class subsets and novel modalities.
As a result, more prompts and the associated trainable parameters are required to learn an expanding set of classes and modalities.
This results in inaccurate knowledge retrieval (Table \ref{tab: ablation study components} in the main paper) and greater memory consumption (Figure \ref{fig: param growth} in the main paper).

To clearly demonstrate the intrinsic maintainability of previous features by self-regularization (\textit{Self}) after learning subsequent tasks, we visualize the embeddings obtained using the proposed method and the method without \textit{Self} in Figure \ref{fig: t-sne}.
We compared the embeddings of tasks extracted right after training each task with those extracted after training all subsequent tasks.
The results show that the embeddings remain almost unchanged even after learning new tasks compared to the previously extracted embeddings, when applying the presented \textit{Self}. 
However, after learning subsequent tasks, embeddings from the method without \textit{Self} become mixed, especially in the video and depth modalities.

\renewcommand{\arraystretch}{0.8}
\begin{table}[t!]
\centering
\scriptsize
\resizebox{\columnwidth}{!}{%
\begin{tabular}{cr|ccccc}
\hline
\rowcolor{gray!15}                    &     & L2P & S-liPrompts & AttriCLIP & PGP & COMM \\ \hline
\multirow{3}{*}{\rotatebox[origin=c]{90}{Image}}
& AIA ($\uparrow$) & 76.9   & 78.4  & 77.3    & 79.5    & \textbf{85.8}     \\
& FAA ($\uparrow$) & 68.6   & 67.5  & 65.7    & 70.2    & \textbf{79.3}     \\
& F   ($\downarrow$) & 10.3   & 13.7  & 14.6    & 11.7    & \textbf{8.1}     \\ \hline
\multirow{3}{*}{\rotatebox[origin=c]{90}{Audio}}
& AIA ($\uparrow$)   & 44.5     & 41.3      & 41.7      & 44.8    & \textbf{61.7}     \\
& FAA ($\uparrow$)   & 25.0     & 19.3      & 23.5      & 26.2    & \textbf{49.0}    \\
& F   ($\downarrow$) & 24.3     & 27.5      & 22.8      & 23.3    & \textbf{15.9}     \\ \hline
\end{tabular}%
}
\caption{Results on unimodal continual learning}
\label{tab:unimodal_CL}
\end{table}

\newpage
\noindent{\textbf{Study on Prompt Configuration.}}
We conducted an additional study on the design choices of the proposed prompting mechanism.
Specifically, we analyze how prompts should be configured and composed to retain previously learned knowledge and to support new modalities without interference.
This study investigates
(1) the location of prompt insertion (language encoder vs. modality encoder),
(2) the strategy for accumulating prompts within each modality over time, and
(3) the mechanism for composing prompts across modalities based on relevance.
The results of this study are summarized in Table \ref{tab:ablation_prompt_config}.

\noindent\textit{(1) Prompt Insertion Location: }
We evaluate the impact of prompt insertion by comparing three variants of our method: COMM-L (inserting prompts only into the language encoder), COMM-V (only into the modality encoder), and COMM-VL (into both encoders), where COMM-VL corresponds to the proposed method.
The results show that COMM-VL achieves the highest performance, indicating that leveraging prompts in both encoders is essential for learning modality-specific features while maintaining alignment with semantic labels.

\noindent\textit{(2) Intra-Modality Accumulation: }
To mitigate forgetting within each modality, we introduce a self-regularized accumulation strategy that incrementally incorporates newly learned prompts while preserving the semantic continuity of previously acquired ones. This approach is compared against two simplified alternatives: element-wise summation (COMM-Sum) and averaging (COMM-Avg) of past prompts. Results indicate that these naïve methods substantially degrade performance, confirming that uncontrolled accumulation disrupts discriminative prompt representations. The self-regularization strategy is thus essential for ensuring temporal consistency during intra-modal knowledge integration.

\noindent\textit{(3) Inter-Modality Composition.}
To assess the effectiveness of the proposed relevance-guided prompt composition across modalities, we compare it with a non-parametric baseline that estimates modality similarity using Mahalanobis distance computed over Gaussian-sampled features. 
While both methods utilize the same statistical feature summary, the learned relevance scores in the proposed approach lead to consistently superior results. 
This indicates that learning relevance scores enables more precise and robust identification of semantically related modalities, resulting in more stable and effective cross-modal prompt integration.

\vspace{1mm}
\noindent{\textbf{Results on unimodal continual learning.}}
While COMM is designed for multimodal continual learning, it can also be applied to unimodal learning scenarios.
To validate this, we conducted additional continual learning experiments where each task consists of a single non-text modality (image or audio) paired with text.
Specifically, we use ImageNet-100 for image-text tasks and ESC-50 for audio-text tasks, splitting each into five incremental tasks.
As shown in Table \ref{tab:unimodal_CL}, COMM significantly outperforms other prompt-based continual learning methods (L2P, S-liPrompts, AttriCLIP, and PGP) in both modalities, achieving higher average accuracy (AIA, FAA) and lower forgetting (F).
This shows that the proposed method not only generalizes to multimodal settings but also improves learning stability in conventional unimodal tasks.

\end{document}